\pdfoutput=1

\documentclass[11pt]{article}

\usepackage{ACL2023}

\usepackage{times}
\usepackage{latexsym}
\usepackage{booktabs}
\usepackage{array}
\usepackage{adjustbox}
\usepackage{tabularx}
\usepackage[T1]{fontenc}
\usepackage{mathtools}

\usepackage[utf8]{inputenc}

\usepackage{microtype}
\usepackage{inconsolata}
\usepackage{graphicx}
\usepackage[ruled,vlined]{algorithm2e}
\usepackage{algorithmic}
\usepackage{amsmath}
\usepackage{amssymb}
\usepackage{changepage}
\usepackage{makecell}
\usepackage{enumitem}
\newcommand{\method}{\textsc{GenICL}\xspace}

\title{Learning to Select In-Context Demonstration Preferred\\by Large Language Model}

\author{
Zheng Zhang$^{1,3}$, Shaocheng Lan$^{1}$, Lei Song$^{2}$\footnotemark[1]$\ $, Jiang Bian$^{2}$, Yexin Li$^{3}$, Kan Ren$^{1}$\thanks{\ \ Corresponding author.} \\
$^{1}$School of Information Science and Technology, ShanghaiTech University \\
$^{2}$Microsoft Research Asia \\
$^{3}$State Key Laboratory of General Artificial Intelligence, BIGAI,
Beijing, China \\
\texttt{\{zhangzheng2024, lanshch2024, renkan\}@shanghaitech.edu.cn} \\
\texttt{\{lei.song, jiang.bian\}@microsoft.com, liyexin@bigai.ai}
}

\begin{document}
\maketitle
\begin{abstract}
In-context learning (ICL) enables large language models (LLMs) to adapt to new tasks during inference using only a few demonstrations. 
However, ICL performance is highly dependent on the selection of these demonstrations. 
Recent work explores retrieval-based methods for selecting query-specific demonstrations, but these approaches often rely on surrogate objectives such as metric learning, failing to directly optimize ICL performance.
Consequently, they struggle to identify truly beneficial demonstrations. 
Moreover, their discriminative retrieval paradigm is ineffective when the candidate pool lacks sufficient high-quality demonstrations.
To address these challenges, we propose \method, a novel generative preference learning framework that leverages LLM feedback to directly optimize demonstration selection for ICL.
Experiments on 19 datasets across 11 task categories demonstrate that \method achieves superior performance than existing methods in selecting the most effective demonstrations, leading to better ICL performance.
The project website is at \url{https://foundation-model-research.github.io/GenICL}.

\end{abstract}

\section{Introduction}
Large Language Models (LLMs) have demonstrated remarkable capabilities across a wide range of tasks. 
In-context learning (ICL), introduced by \citep{brown2020language}, enables LLMs to perform tasks with only a few examples as demonstration, without requiring parameter updates \citep{brown2020language,liu2021makes}.

Although ICL is promising, the selection of in-context demonstrations is crucial, as it significantly impacts LLM performance \citep{lu2021fantastically,min2022rethinking}.
Many existing approaches rely on retrieval-based methods, either using off-the-shelf retrievers \citep{robertson2009probabilistic} or training specialized ones \citep{reimers2019sentence,wang2022text}. 
Training-based methods typically optimize retrievers to approximate the relevance score of the candidate to the given query for downstream ICL tasks, often leveraging metric learning-based proxy tasks, such as contrastive loss \citep{rubin2021learning,ye2023compositional,wang2023learning,cheng2023uprise,luo2023dr,liu2024demorank}.

\begin{figure}
    \centering  
    \includegraphics[width=\columnwidth]{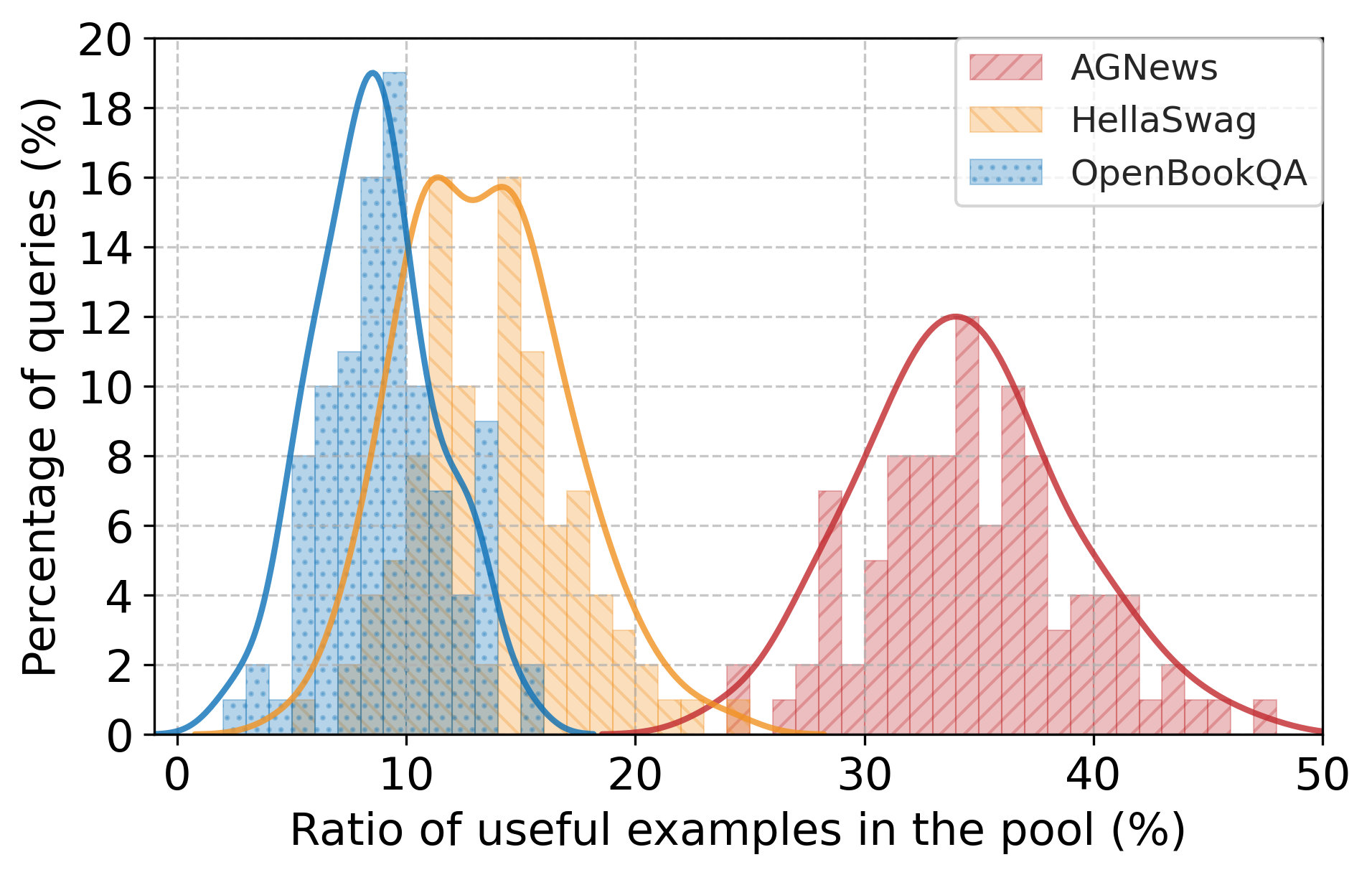} 
    \caption{
    The distribution of the useful example ratio across test sets from different datasets.
    The x-axis represents the ratio $\frac{\text{\# useful examples}}{\text{\# total examples}}$, where a `useful example' is one that helps the LLM generate an accurate output with in-context learning.
    The ratio remains low for most test queries, i.e., the majority of demonstration examples are ineffective for the LLM.
    }
    \label{fig:acc_distribution}  
\end{figure}

However, these methods face several challenges. 
The most critical issue is the misalignment between the surrogate learning objective of the retriever and the intrinsic optimization goal of ICL. 
Specifically, a discriminative model, trained with a metric learning objective to approximate the relevance score between a demonstration candidate and a given query, does not necessarily indicate the candidate's effectiveness as an in-context demonstration for an LLM.
Furthermore, the scarcity of effective demonstration candidates in the retrieval pool poses another challenge for training a discriminative retriever. 
Figure~\ref{fig:acc_distribution} illustrates the distribution of test queries regarding the ratio of useful examples that help the LLM achieve accurate output using ICL.
The figure shows that most candidates are ineffective for most queries, making retriever optimization particularly challenging.
This issue is further exacerbated by the properties of hard labels for LLM feedback, as noted by \citep{wang2023learning}.

To address these challenges, we propose \method, a novel generative preference learning framework that directly optimizes demonstration selection for ICL using LLM feedback.
Specifically, we reformulate ICL as a generative Bayesian optimization problem, introducing a latent variable to bridge demonstration selection and LLM inference.
\method then optimizes this objective through preference learning, which focuses more on the relative effectiveness between demonstration samples, allowing it to capture finer-grained information in scenarios where effective demonstrations are scarce.
This optimization procedure on demonstration selection more closely aligns with the intrinsic objective of ICL.
Through this approach, \method can better capture the LLM's demonstration preference and identify the truly effective demonstrations (preferred by the LLM).
To validate the effectiveness of \method, we conducted experiments on a diverse set of datasets covering classification, multiple-choice, and generation tasks. 
The results demonstrate that \method achieves superior performance compared to existing methods.

Our contributions can be summarized as follows:
\begin{itemize}[leftmargin=3mm]
  \item We propose \method, a novel generative preference learning framework that directly optimizes demonstration selection for ICL using LLM feedback, overcoming the limitations of sub-optimality of surrogate learning objective used in traditional retrieval-based methods.
  \item The preference learning of paired effective and ineffective demonstrations leads to fine-grained distinguishness of demonstration preference of LLM.
  \item We conduct extensive experiments on 19 datasets across 11 task categories with a quantitative and qualitative analysis of the effectiveness of our proposed method.
\end{itemize}

\section{Related work}\label{sec:related-work}

\subsection{In-context Demonstration Selection}
In-context learning \citep{brown2020language} enhances large language models by leveraging few-shot examples for task adaptation. 
While proficient at using provided examples, LLMs heavily depend on specific demonstrations \citep{min-etal-2022-metaicl,luo2024context}, making demonstration selection crucial for downstream performance.
Existing approaches to demonstration selection can be broadly categorized into two branches. 
The first employs off-the-shelf retrievers like BM25 \citep{robertson2009probabilistic}, SBERT \citep{reimers2019sentence}, and E5\textsubscript{base} \citep{wang2022text}, which prioritize semantic similarity but overlook downstream utility. 
The second category involves learning the retriever based on feedback from LLM performance on selected demonstrations \citep{li2023unified,cheng2023uprise,luo2023dr,chen2024learning}.
For instance, \citet{rubin2021learning} trained a dense retriever using LLM-generated feedback, while \citet{wang2023learning} distilled a retriever from a metric function estimating relevance scores between demonstration candidates and the query. 
However, these methods approximate the relevant score of the candidate given the query, which is not essentially aligned with the true optimization objective of ICL resulting in a sub-optimal solution.

Moreover, only a minority of candidates truly enhance in-context learning performance (Figure \ref{fig:acc_distribution}), making it challenging to learn a reasonably dense retriever.

Therefore, a more promising method is required for directly optimizing the intrinsic ICL objective.

\subsection{Preference Learning}
Preference learning plays a vital role in training LLMs to align generative models with human values and preferences \citep{christiano2017deep, ziegler2019fine, ouyang2022training}. 
It has also emerged as a powerful paradigm for enhancing LLM performance in complex tasks \citep{havrilla2024teaching, li2024boosting}.
A key approach in this domain is reinforcement learning from human feedback (RLHF) \citep{ouyang2022training}, which involves training a reward model to capture human preferences and optimizing the model using reinforcement learning algorithms \citep{schulman2017proximal}. 
More recent methods, such as Direct Preference Optimization (DPO) \citep{rafailov2024direct} and Kahneman-Tversky Optimization (KTO) \citep{ethayarajh2024kto}, bypass explicit reward modeling and instead learn preferences directly from human feedback.
While these methods have advanced preference learning performance, they primarily focus on human feedback without explicitly addressing the issue of the scarcity of effective demonstrations.

\begin{figure*}[t]
  \centering
  \includegraphics[width=\textwidth]{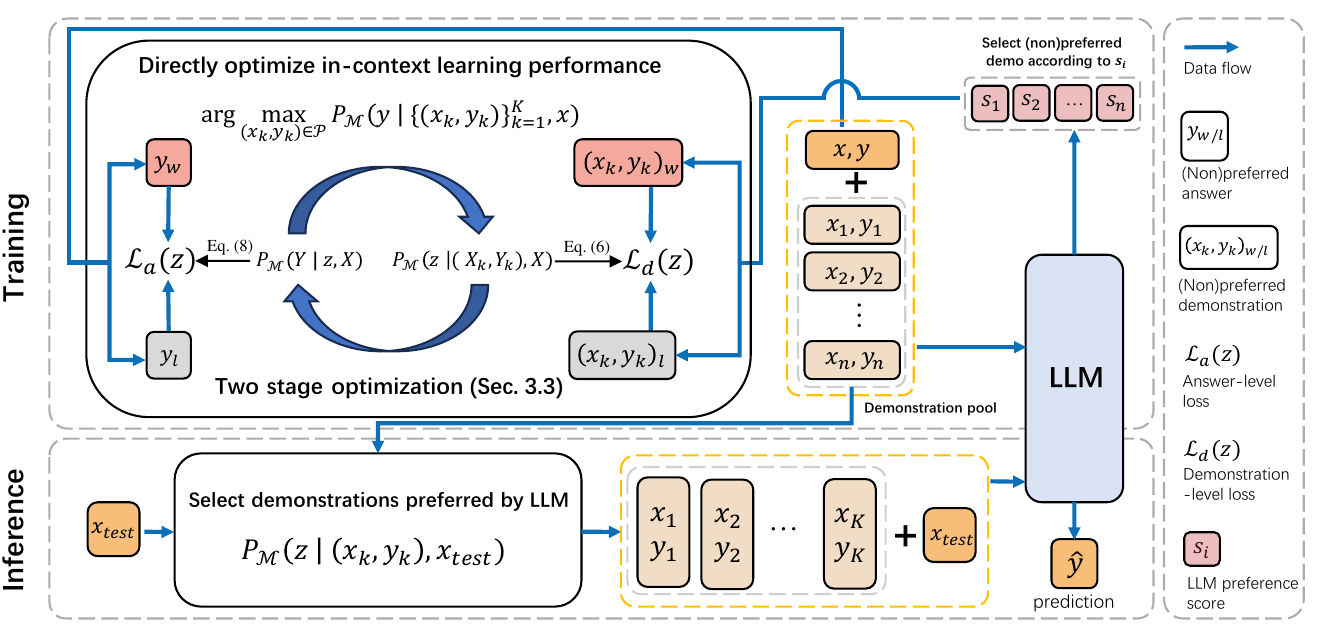} 
  \caption{Training and inference pipeline of \method.}
  \label{fig:pipeline}
\end{figure*}

\section{Methodology}\label{sec:method}
\subsection{Problem Setting}
Consider a task sample $x$ with ground-truth $y$, denoted as $(x,y)$, and a set of demonstrations $\left\{ (x_k, y_k) \right\}_{k=1}^K \subseteq \mathcal{P}$ where each demonstration $(x_k,y_k)$ is an input-output tuple taken from the demonstration pool $\mathcal{P}$ containing all training samples from all tasks.
$K$ is the number of demonstrations used for the current query. 
Our goal is to provide the appropriate demonstrations for each $x$ to improve the ICL performance of LLMs as
\begin{equation} \label{ICL}
\underset{(x_k, y_k) \in \mathcal{P}}{\arg\max} P_{\mathcal{M}} (y \mid \left\{ (x_k, y_k) \right\}_{k=1}^K, x)
\end{equation}

\subsection{Optimization Objective}

Most methods based on LLM feedback used metric learning, such as contrastive learning, as a surrogate objective to train a discriminative model as the retriever. However, this optimization objective has a gap with in-context learning performance (details in Appendix \ref{metric learning}). 
Additionally, the relevance scores determined by the discriminative retriever come from a feature similarity perspective, overlooking optimizations at the semantic understanding level, which leads to the selected demonstrations that may not be truly useful for improving the LLM's in-context learning performance.
To address these issues, we propose a generative framework to directly optimize demonstration selection for ICL performance, thereby avoiding the use of surrogate objectives that have a gap. We treat in-context learning as generative Bayesian optimization and introduce a latent variable $z$ to establish the relationship between demonstration selection and ICL performance.
The optimization problem is formulated as follows:
\begin{multline} \label{wang equation}
P_{\mathcal{M}}(Y \mid \{(X_k, Y_k)\}_{k=1}^K, X) = \\
\int_{z} P_{\mathcal{M}}(Y \mid z, X)\, P_{\mathcal{M}}(z \mid \{(X_k, Y_k)\}_{k=1}^K, X)\, dz
\end{multline}
This transformation is also supported by the perspectives of \citet{xie2021explanation} and \citet{wang2024large}, where in-context learning can be viewed as a Bayesian inference process.

The latent variable $z$ is viewed as the representation of the LLM's demonstration preference. 
Through $z$, we concretize the LLM's preference for demonstration and use it to identify effective demonstrations. Eq. (\ref{wang equation}) proves from a Bayesian perspective that, during the optimization process, both the ground truth and demonstrations should be considered simultaneously. 
However, in the surrogate optimization objectives of retriever-based methods, only the quality of the demonstrations is considered. This is why demonstrations selected by these methods do not necessarily enhance the LLM's ability to predict the ground truth within ICL.

To obtain the demonstration preference, we propose to fully optimize the latent variable \(z\) by jointly considering the ground truth and the preference of the demonstrations. 
We utilize the Evidence Lower Bound (ELBO) \citep{kingma2013auto} and Eq. (\ref{wang equation}) to obtain the following optimization objective, with the detailed derivation provided in Appendix \ref{Derivation of the Optimization Objective}.
\begin{multline}
\small
\mathcal{L}(z) = -\log P_{\mathcal{M}}(Y \mid z, X) \\
- \log P_{\mathcal{M}}(z \mid \{(X_k, Y_k)\}_{k=1}^K, X)
\end{multline}
Due to the combinatory search space of $(X_1,Y_1),(X_2,Y_2),...,(X_K,Y_K)$ being immense, we simplify the optimization objective to the following formula.
\begin{multline}\label{eq:icl-loss}
\mathcal{L}(z) = -\log P_{\mathcal{M}}(Y \mid z, X) \\
-\log P_{\mathcal{M}}(z \mid (X_k, Y_k), X)
\end{multline}

\subsection{Demonstration Preference Learning}
The above optimization objective aims to find the variable $z$ of the demonstration preference of the LLM. From an implementation perspective, $z$ is a task-specific description, and its corresponding trainable parameters are $\theta$. We optimize $\theta$ to refine the demonstration preference distribution corresponding to $z$.  

To perform effective optimization in scenarios where effective demonstrations are scarce, we use preference learning, as it focuses more on the relative relationships between demonstrations, enabling it to capture finer-grained information. To learn the preference for effective demonstrations, we adopt the KTO algorithm \citep{ethayarajh2024kto}, which directly maximizes the utility of LLM generations based on context information, conditioned on either the effective demonstration $(x_k,y_k)_w$ or the relatively ineffective one $(x_k, y_k)_l$. This ensures that $(x_k,y_k)_w \succ (x_k,y_k)_l$ given the input $x$. To achieve this, we reformulate the optimization objectives, $-\log P_{\mathcal{M}}(Y \mid z, X)$ and $-\log P_{\mathcal{M}}(z \mid (X_k, Y_k), X)$ in Eq.~\eqref{eq:icl-loss}, as a two-stage alternating optimization process.

\paragraph{The First Stage.}
In this stage, we optimize the demonstration selection part $-\log P_{\mathcal{M}}(z \mid (X_k, Y_k), X)$ in Eq.~\eqref{eq:icl-loss} with preference learning.
Preference learning requires both preferred data and non-preferred data.
To obtain the LLM preferences for demonstrations under the in-context learning paradigm, we utilize the ICL performance of LLM given each demonstration candidate as the preference score.

The preference score $s_k$ is the log-likelihood of LLM generation w.r.t. the ground-truth output $y$ given the candidate $(x_k, y_k)$ and the input $x$, reflecting the LLM's preference for the demonstration.

\begin{equation} \label{preference_score}
s_k = P(y|(x_k,y_k), x)
\end{equation}

Since we need to score all the training data based on the above method to obtain the preference dataset, and the demonstration pool for each training sample is $\mathcal{P}$, the complexity of scoring all the data is quadratic in $|\mathcal{P}|$, which is computationally infeasible for an exhaustive search. 
Following prior work \citep{wang2024large, cheng2023uprise, wang2024effective}, we use E5\textsubscript{base} \citep{wang2022text} to obtain a more relevant subset $\xi \subseteq \mathcal{P}$  as the new demonstration pool to reduce search space. 
For each training sample, we select the set of candidates with the highest preference scores as preferred demonstrations $(x_k, y_k)_w$, and the set of candidates with the lowest preference scores as non-preferred demonstrations $(x_k, y_k)_l$.
The demonstration-level loss is formulated as follows.
\begin{multline} \label{Ld_loss}
\mathcal{L}_{d}(\theta) = 
- \mathbb{E}_{((x_k, y_k)_{w}, (x_k, y_k)_{l}, x, z) \sim \mathcal{D}} \\
\left[ \lambda_w \sigma\left(\beta \log \frac{P_{\mathcal{M}}(z \mid (x_k, y_k)_{w}, x)}{P_{\mathcal{M}_{\text{ref}}}(z \mid (x_k, y_k)_{w}, x)} - s^{d}_{\text{ref}}\right) \right. \\
\left. + \lambda_l \sigma\left(s^{d}_{\text{ref}} - \beta \log \frac{P_{\mathcal{M}}(z \mid (x_k, y_k)_{l}, x)}{P_{\mathcal{M}_{\text{ref}}}(z \mid (x_k, y_k)_{l}, x)}\right) \right]
\end{multline}
\begin{multline}
s^{d}_{\text{ref}} = \mathbb{E}_{(x, y) \sim \mathcal{D}} \biggl[ 
\beta \, \text{KL} \bigl( P_{\mathcal{M}} (z \mid (x_k, y_k), x) \, \\
\parallel \, P_{\mathcal{M}_{\text{ref}}}(z \mid (x_k, y_k), x) \bigr) \biggr]
\end{multline}
where $\beta$ is a hyperparameter, $\lambda_w$ and $\lambda_l$ are hyperparameters for preferred data and non-preferred data respectively. $s_{\text{ref}}$ represents the use of KL divergence as a constraint to prevent the model from deviating too much from its original predictive ability, thereby enhancing the stability of the model.

\paragraph{The Second Stage.} 
We then optimize the model generation part $-\log P_{\mathcal{M}}(Y \mid z, X)$ in Eq.~\eqref{eq:icl-loss} in a similar manner of preference learning.
For all tasks, we treat the ground truth as the correct answer $y_w$. For classification and multi-choice tasks, we consider the incorrect categories or options as wrong answers $y_l$, and for generation tasks, we randomly sample answers from other samples as $y_l$ for the current sample. The answer-level loss is formulated as follows:
\begin{multline} \label{La_loss}
\mathcal{L}_{a}(\theta) = 
- \mathbb{E}_{(z, x, y_w, y_l) \sim \mathcal{D}} \\
\left[ \lambda_w \sigma\left(\beta \log \frac{P_{\mathcal{M}}(y_w \mid z, x)}{P_{\mathcal{M}_{\text{ref}}}(y_w \mid z, x)} - s^{a}_{\text{ref}}\right) \right. \\
\left. + \lambda_l \sigma\left(s^{a}_{\text{ref}} - \beta \log \frac{P_{\mathcal{M}}(y_l \mid z, x)}{P_{\mathcal{M}_{\text{ref}}}(y_l \mid z, x)}\right) \right] 
\end{multline}

\begin{multline}
s^{a}_{\text{ref}} = \mathbb{E}_{(x, y) \sim \mathcal{D}} \biggl[ 
\beta \, \text{KL} \bigl( P_{\mathcal{M}} (y \mid z, x) \, \\
\parallel \, P_{\mathcal{M}_{\text{ref}}}(y \mid z, x) \bigr) \biggr]
\end{multline}
where the hyperparameters are consistent with those in the first stage.

In summary, we alternately optimize the losses of the two stages, and the complete optimization objective is as follows:
\begin{equation}
\mathcal{L}(\theta) = \mathcal{L}_{a}(\theta) + \mathcal{L}_{d}(\theta)
\end{equation}
And Algorithm \ref{Demonstration Preference Learning} summarizes the whole training procedure.

\begin{algorithm}[htbp]
\caption{Demonstration Preference Learning}
\label{Demonstration Preference Learning}
\begin{adjustwidth}{-1em}{0pt}
\begin{algorithmic}[1]
    \STATE \textbf{Input:} Task dataset $\mathcal{D} = \{(x, y)_i\}_{i=1}^{|\mathcal{D}|}$, latent variable $z$, demonstration pool $\mathcal{P}$, and learning rates $\eta_1, \eta_2$
    \STATE \textbf{Initialization}: Parameters $\theta$
    \FOR{each $(x, y)$ in task $\mathcal{D}$}
        \STATE Calculate demonstration scores by Eq. (\ref{preference_score}), and identify preferred demonstrations $(x_k, y_k)_w$ and non-preferred ones $(x_k, y_k)_l$
        \STATE Compute demonstration-level loss $\mathcal{L}_{d}(\theta)$ by Eq. (\ref{Ld_loss}) 
        \STATE Update $\theta \gets \theta - \eta_1 \nabla_\theta \mathcal{L}_d(\theta)$
        \STATE Set the correct answer as $y_w = y$, and select an incorrect answer $y_l$
        \STATE Compute answer-level loss $\mathcal{L}_{a}(\theta)$ by Eq. (\ref{La_loss}) 
        \STATE Update $\theta \gets \theta - \eta_2 \nabla_\theta \mathcal{L}_a(\theta)$
    \ENDFOR
\end{algorithmic}
\end{adjustwidth}
\end{algorithm}

\subsection{Demonstration Selection in Inference}
In the training stage, we optimize the parameters $\theta$ to obtain the variable $z$ of the demonstration preference of the LLM.
In the inference stage, we first use E5\textsubscript{base} \citep{wang2022text} to reduce the search space and time cost of demonstration selection. 
Then we leverage the LLM with the latent variable $z$ to select demonstrations for an input query $x_{\text{test}}$. In Eq. (\ref{wang equation}), for the same $x$, all demonstrations have the same $P_{\mathcal{M}}(y \mid z, x)$. 
Therefore, we select the top-$K$ demonstrations based on the probability of generating the latent variable, as shown below:
\begin{equation} \label{Demonstration Selection}
P_{\mathcal{M}}(z \mid (x_k, y_k), x_{\text{test}})   
\end{equation}
After obtaining a set of demonstrations $\{(X_k, Y_k)\}_{k=1}^K$, we use the frozen LLM to perform in-context learning.
The entire training and inference pipeline is shown in Figure \ref{fig:pipeline}.

\section{Experiments}\label{sec:exp}

\subsection{Evaluations Setup}

\paragraph{Datasets.}
    We validate the effectiveness of \method across a range of language processing tasks including classification, multi-choice, and text generation. 
    The classification tasks include \textbf{Natural Language Inference} (RTE \citep{wang2018glue}, SNLI \citep{bowman2015large}), \textbf{Paraphrase} (PAWS \citep{zhang2019paws}, QQP \citep{wang2018glue}), \textbf{Reading Comprehension} (MultiRC \citep{khashabi2018looking}, BoolQ \citep{clark2019boolq})), \textbf{Sentiment Classification}: (SST2 \citep{socher2013recursive}, Sentiment140 \citep{go2009twitter}), and \textbf{Topic Classification} (AGNews \citep{zhang2015character}).
    The multi-choice tasks cover \textbf{Commonsense Reasoning} (COPA \citep{roemmele2011choice}, HellaSwag \citep{zellers2019hellaswag}, OpenBookQA \citep{mihaylov2018can}), \textbf{Coreference} (Winogrande \citep{sakaguchi2021winogrande}).
    The text generation tasks contain \textbf{Close QA} (NaturalQuestions \citep{kwiatkowski2019natural}, SQuAD v1 \citep{rajpurkar2016squad}), \textbf{Commonsense Generation} (CommonGen \citep{lin2019commongen}), \textbf{Data-to-Text} (E2E NLG \citep{duvsek2019semantic}, DART \citep{nan2020dart}), and \textbf{Summarization} (AESLC \citep{zhang2019email}, Gigaword \citep{napoles2012annotated}).

\paragraph{Baseline.} 
Our evaluation compares \method against several established baseline approaches that leverage the ICL capabilities of LLMs. These baselines include zero-shot (which executes tasks without demonstrations), random (which selects demonstrations arbitrarily from the available pool), and off-the-shelf retrieval models such as BM25~\citep{robertson2009probabilistic}, SBERT~\citep{reimers2019sentence}, and E5\textsubscript{base}~\citep{wang2022text}. Furthermore, we conduct comparative analyses against the following more advanced approaches.
\begin{itemize}[leftmargin=3mm]
    \item EPR \citep{rubin2021learning} is a prominent method using LLM feedback to label candidate examples as positive or negative, to train the retriever.
    \item LLM-R \citep{wang2023learning} trains the retriever using a reward model and knowledge distillation, which is one of the state-of-the-art retriever-based methods.
    \item CBDS (concept-based-demonstration-selection) \citep{wang2024large} first learns the latent variable and then selects demonstrations with the highest probability of predicting the corresponding latent variable.
\end{itemize}

\paragraph{Implementation Details.} 
To ensure a fair comparison with the baselines, we use the same template and evaluation metrics. For detailed information, please refer to the Appendix \ref{Templates}. Our approach demonstrates enhanced efficiency through a reduced number of training parameters relative to retriever-based methodologies such as LLM-R~\citep{wang2023learning} and E5\textsubscript{base}~\citep{wang2022text}.
For model implementation, we utilize Llama-7B \citep{touvron2023llama} as our primary LLM for most experiments. We also evaluate the impact and generalizability of our \method in Section \ref{sec:Validation_on_More_LLMs} using GPT-Neo 2.7B \citep{black2021gpt}, Vicuna-13B \citep{zheng2023judging}, LLaMA-3.2-3B \citep{grattafiori2024llama}, and Qwen2.5-3B \citep{yang2024qwen2}.
Our inference protocol incorporates a set of 8 demonstrations within the context, with complete implementation details provided in Appendix \ref{Implementation Details}.

\subsection{Main Results}
Table~\ref{Main Results} presents the performance comparison of the proposed \method and various baselines for classification, multi-choice, and text generation tasks. The results demonstrate that \method significantly outperforms all the baselines on most tasks. 
Zero-shot and Random consistently yield the worst performance across most tasks, highlighting the importance of using high-quality demonstrations in ICL.
Retriever methods trained based on LLM feedback provide only limited improvements over off-the-shelf retrievers. Specifically, EPR performs worse than E5\textsubscript{base} on tasks such as MultiRC, RTE, and Sentiment140. 
This indicates that these discriminative methods, which use proxy optimization objectives, often fail to select truly useful demonstrations. Our generative Bayesian method \method, in contrast, significantly outperforms EPR and LLM-R, proving its effectiveness.
Although \method uses E5\textsubscript{base} to narrow the search space, compared to directly using E5\textsubscript{base}, we achieve significant improvements, especially on tasks such as PAWS, BoolQ, and SNLI, where our \method improves by more than 5 points. 
This indicates that directly using E5\textsubscript{base} struggles to identify demonstrations preferred by the LLM, whereas our method effectively addresses this issue.
While CBDS employs a comparable latent concept variable methodology for demonstration selection, its performance is constrained due to adopting a misaligned optimization objective (details in Appendix \ref{Implementation Details}). Through the implementation of a more refined optimization objective, \method demonstrates superior performance.

For text generation tasks, the improvements of EPR and LLM-R over off-the-shelf retrievers are all relatively limited. This may be because metrics like ROUGE and EM typically exhibit a narrower range of variation compared to classification accuracy, which is consistent with the findings of \citet{wang2023learning}. Even so, \method still achieves significant improvements over LLM-R and off-the-shelf retrievers, particularly on the CommonGen and SQuAD v1 tasks. This indicates that \method is also highly effective for text generation tasks.
We also provide a case study analysis in Appendix \ref{appendix:case_study} to offer further insights.

\begin{table*}[h]
    \centering
    \resizebox{1.0\textwidth}{!}{
    \begin{tabular}{lccccccccc} 
        \toprule
        & \multicolumn{9}{c}{\textbf{Classification Tasks}} \\ 
        \cmidrule(lr){2-10}
        & \multicolumn{1}{c}{\textbf{Topic}} & \multicolumn{2}{c}{\textbf{Reading Comprehension}} & \multicolumn{2}{c}{\textbf{Paraphrase}} & \multicolumn{2}{c}{\textbf{NLI}} & \multicolumn{2}{c}{\textbf{Sentiment}} \\ 
        \cmidrule(lr){2-2} \cmidrule(lr){3-4} \cmidrule(lr){5-6} \cmidrule(lr){7-8} \cmidrule(lr){9-10}
        & AGNews & BoolQ & MultiRC  & PAWS & QQP & RTE & SNLI & \makecell{Sentim\\ent140} & SST2 \\ 
        \midrule
        Zero-shot  & 31.4  & 64.7  & 57.0  & 53.0  & 57.9 & 59.9 & 39.6 & 49.3 & 54.2    \\ 
        Random     & 65.0  & 69.6  & \textbf{60.4}   & 49.6  & 54.0 & 65.7 & 40.4 & 78.8 & 64.1    \\ 
        BM25       & 90.0  & 74.0  & 58.7   & 56.5  & 80.3 & 59.9 & 47.7 & 88.3 & 84.7    \\ 
        SBERT      & 89.8  & 73.6  & 53.3  & \underline{58.3}  & \underline{81.7} & 60.2 & 56.2 & \underline{94.1} & 87.8    \\ 
        E5\textsubscript{base}     & 90.6  & 71.0  & 54.0  & 55.6  & 77.3 & \underline{68.5} & 53.7 & 93.0 & 92.4    \\ 
        CBDS       & 67.3  & \underline{77.6}  & 49.3    & 57.6  & 64.2 & 56.3 & 43.5 & 92.5 & 69.2   \\ 
        EPR        & 91.8  & 74.8  & 50.4   & 57.7  & \underline{81.7} & 66.8 & 68.4 & 91.4 & 88.7   \\ 
        LLM-R      & \underline{92.4}  & 74.9  & 50.2  & 57.5  & 80.9 & 61.7 & \underline{80.0} & 91.6 & \underline{93.4}  \\ 
        \method (ours)   & \textbf{92.6}  & \textbf{78.1}  & \underline{56.9}   & \textbf{63.9}  & \textbf{82.0} & \textbf{72.9} & \textbf{84.6} & \textbf{94.7} & \textbf{95.0}  \\ 
        \bottomrule
    \end{tabular}
    }
    
    \resizebox{1.0\textwidth}{!}{
    \begin{tabular}{lcccccccccc} 
        \toprule
        & \multicolumn{4}{c}{\textbf{Multi-Choice Tasks}} & \multicolumn{6}{c}{\textbf{Text Generation Tasks}} \\ 
        \cmidrule(lr){2-5} \cmidrule(lr){6-11}
        & \multicolumn{1}{c}{\textbf{\makecell{Corefe\\rence}}} & \multicolumn{3}{c}{\textbf{\makecell{Commonsense \\Reasoning}}} & \multicolumn{2}{c}{\textbf{Summarize}} & \multicolumn{1}{c}{\textbf{\makecell{Comm\\onGen}}} & \multicolumn{2}{c}{\textbf{Data-to-text}} & \multicolumn{1}{c}{\textbf{CloseQA}} \\ 
        \cmidrule(lr){2-2} \cmidrule(lr){3-5} \cmidrule(lr){6-7} \cmidrule(lr){8-8} \cmidrule(lr){9-10} \cmidrule(lr){11-11}
        & \makecell{Winog\\rande} & COPA & \makecell{Hella\\Swag} & \makecell{Open\\BookQA} & \makecell{AE\\SLC} & \makecell{Giga\\word} & \makecell{Comm\\onGen} & DART & \makecell{E2E\\NLG} & \makecell{SQuAD\\v1} \\ 
        \midrule
        Zero-shot & 61.8 & 66.0 & 71.5 & 41.6 & 5.7  & 15.4  & 19.2  & 22.8  & 34.6  & 2.2  \\ 
        Random    & 62.1 & 74.0 & 72.1 & 43.0 & 6.5  & 27.2  & 36.3  & 34.5  & 51.1  & 47.3  \\ 
        BM25      & 66.4 & 77.0 & 74.6 & 47.8 & 23.9 & \underline{32.5}  & 38.3  & 55.4  & 53.8  & 53.7  \\ 
        SBERT     & \underline{66.9} & 81.0 & \textbf{75.2} & 49.2 & 22.3 & 31.7  & 37.8  & 54.6  & 50.6  & 62.5  \\ 
        E5\textsubscript{base} & 66.7 & 84.0 & 75.0 & \underline{51.0} & 23.4 & 31.9  & 37.6  & 54.9  & 52.3  & 61.9  \\ 
        CBDS      & 66.3 & 83.0 & 73.6 & 47.6 & 20.5 & 29.2  & 34.5  & 52.2  & 50.8  & 59.8  \\ 
        EPR       & 66.5 & 82.0 & \textbf{75.2} & 49.6 & \textbf{26.0} & 32.4  & \underline{39.2}  & \underline{56.2}  & 53.6  & \underline{64.3}  \\ 
        LLM-R     & \underline{66.9} & \underline{85.0} & 74.6 & 50.8 & \textbf{26.0} &  \underline{32.5}  & 37.2  & 56.0  & \underline{54.4}  & 61.8  \\ 
        \method (ours)  & \textbf{68.0} & \textbf{86.0} & 74.6 & \textbf{51.8} & 24.4 & \textbf{33.0}  & \textbf{41.0}  & \textbf{56.4}  & \textbf{55.1}  & \textbf{65.7}  \\
        \bottomrule
    \end{tabular}
    }
    \caption{Main results on classification, multi-choice, and text generation tasks.}
    \label{Main Results}
\end{table*}

\subsection{Ablation Study}
To validate the effectiveness of the key design in \method, we conduct ablation studies on the CommonGen, BoolQ, and SQuAD v1 tasks, with results presented in Table~\ref{Ablation Study}. 
When non-preferred data is removed during training, the performance of \method drops significantly, indicating that using only useful demonstrations without considering relative preference relations makes it difficult to learn the distribution of LLM demonstration preference. This proves the necessity of optimization based on preference learning.
Our complete optimization objective consists of two parts: answer-level loss and demonstration-level loss. Removing either leads to a performance drop, demonstrating the correctness of our proposed optimization framework and the importance of considering both the ground truth and the quality of demonstrations during optimization.

\begin{table}[h]
    \centering
    \small
    \begin{tabular}{l@{\hspace{-5mm}}rccc}
        \toprule
        & CommonGen & BoolQ & SQuAD v1 \\ 
        \midrule
        \method                        & 41.0 & 78.1 & 65.7 \\ 
        - w/o non-preferred data    & 38.7 & 71.2 & 62.5 \\ 
        - w/o answer-level loss     & 34.9 & 72.7 & 62.0 \\ 
        - w/o demo-level loss & 37.9 & 65.8 & 61.7 \\ 
        \bottomrule
    \end{tabular}
    \caption{Ablation study.}
    \label{Ablation Study}
\end{table}

\subsection{Generalizability and Scalability}
\label{sec:Validation_on_More_LLMs}
\paragraph{Validation on More LLMs.}
The previous experimental results were obtained using the LLaMA-7B model. To comprehensively assess the effectiveness and generalizability of \method, we expanded our evaluation to encompass various LLMs of different scales, specifically GPT-Neo 2.7B, Vicuna-13B, LLaMA-3.2-3B, and Qwen2.5-3B. The comparison results are presented in Table~\ref{Other LLMs}. For each model, we report the performance of \method alongside zero-shot setting and a strong baseline, LLM-R \citep{wang2023learning}. Across all models and tasks, \method consistently achieves the best results, suggesting that its benefits are not limited to high-capacity LLMs. For example, with the relatively smaller GPT-Neo 2.7B, \method still leads with 60.5 on PAWS and 92.7 on SST2 task. This indicates that \method can compensate for limited model capacity by better utilizing demonstration. These results provide compelling evidence that \method is a model-agnostic and scalable method.

\begin{table}[h]
    \centering
    \small
    \begin{tabular}{l@{\hspace{-5mm}}rcccc}
        \toprule
        & CommonGen & E2E NLG & PAWS & SST2  \\ 
        \midrule
        GPT-Neo 2.7B   \\ 
        - Zero-shot         & 14.2  & 7.71 & 52.6 & 53.7  \\ 
        - LLM-R             & 34.5  & 49.2 & 55.5 & 92.6  \\ 
        - \method (ours)    & 35.3  & 50.1 & 60.5 & 92.7  \\     
        \midrule
        Vicuna-13B   \\ 
        - Zero-shot         & 17.4  & 1.71 & 57.2 & 49.0  \\ 
        - LLM-R             & 39.7  & 55.4 & 64.4 & 93.8  \\ 
        - \method (ours)    & 39.7  & 56.3 & 71.7 & 94.0  \\ 

        \midrule
        Qwen2.5-3B   \\ 
        - Zero-shot         & 23.6  & 43.6 & 58.8 & 51.0  \\ 
        - LLM-R             & 35.6  & 37.0 & 70.4 & 94.3  \\ 
        - \method (ours)    & 37.4  & 37.9 & 81.8 & 94.6  \\  
        \midrule
        LLaMA-3.2-3B   \\ 
        - Zero-shot         & 16.2  & 33.5 & 54.1 & 52.4  \\ 
        - LLM-R             & 39.1  & 56.4 & 56.0 & 93.0  \\ 
        - \method (ours)    & 39.8  & 57.5 & 66.2 & 94.5  \\ 
        \bottomrule
    \end{tabular}
    \caption{Performance on more LLMs.}
    \label{Other LLMs}
\end{table}

\paragraph{Generalizability of Selected Demonstrations.}
In Table \ref{Table: Generalizability}, we conduct experiments to evaluate th generalizability of demonstrations selected by our \method. Specifically, we examine whether demonstrations selected using a smaller model, GPT-Neo 2.7B, can be effectively reused by larger models such as LLaMA-7B and LLaMA-3.2-3B. The results indicate that demonstrations chosen by the smaller model can significantly improve the ICL performance of larger models, without the need for retraining or further adaptation. This highlights the potential of our \method to serve as a lightweight and model-agnostic demonstration selector.

\begin{table}[h]
    \centering
    \small
    \begin{tabular}{l@{\hspace{-5mm}}rcccc}
        \toprule
        & CommonGen & E2E NLG & PAWS & SST2  \\ 
        \midrule
        GPT-Neo 2.7B        & 35.3 & 50.1 & 60.5 & 92.7 \\
        LLaMA-7B            & 40.8 & 55.3 & 59.3 & 94.0 \\
        LLaMA-3.2-3B        & 39.4 & 57.2 & 59.5 & 93.4 \\ 
        \bottomrule
    \end{tabular}
    \caption{Generalizability of Selected Demonstrations. All demonstrations are selected by GPT-Neo 2.7B and used as in-context demonstrations for all three models.}
    \label{Table: Generalizability}
\end{table}

\subsection{The Superiority of Our Optimization Objective}
\begin{figure}[h!]
  \centering
  \includegraphics[width=0.5\textwidth]{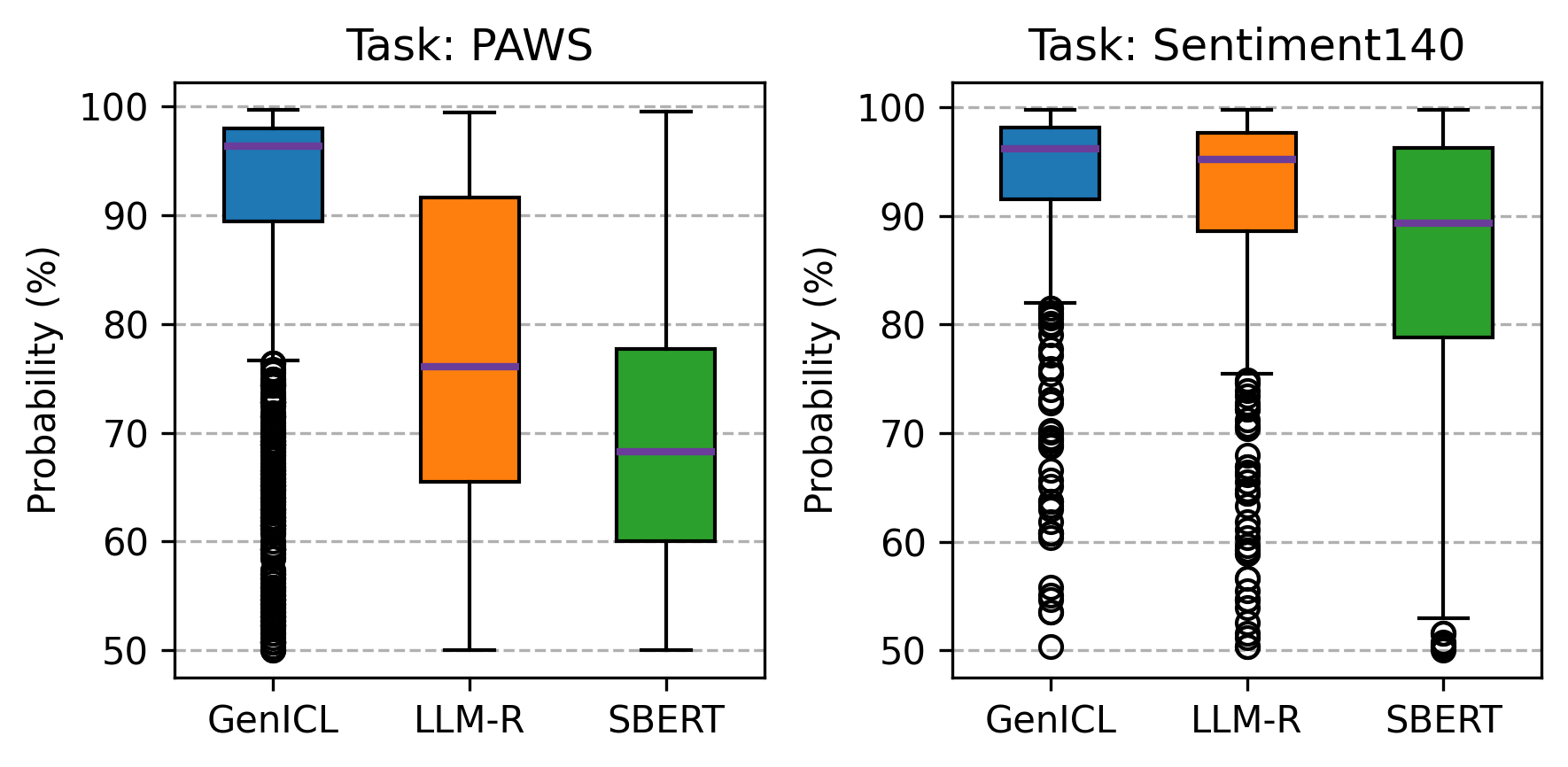} 
  \caption{The predicted probability of the ground truth using demonstrations selected by different methods.}
  \label{box_plot}
\end{figure}

Our optimization objective is directly derived from the ICL paradigm, making it more aligned with ICL compared to retriever-based methods. 
We compare the probability of predicting the ground truth using demonstrations selected by different methods, as described in Eq. (\ref{ICL}).
The results, shown in Figure \ref{box_plot}, indicate that compared to retriever-based methods such as LLM-R and SBERT, \method achieves higher predicted probabilities for the ground truth, with a more concentrated probability distribution. 
This demonstrates that our optimization objective better enhances ICL performance, further validating our previous conclusions.

\subsection{The Impact of the Number and Order of Demonstrations}
\begin{figure}[h!]
  \centering
  \includegraphics[width=0.5\textwidth]{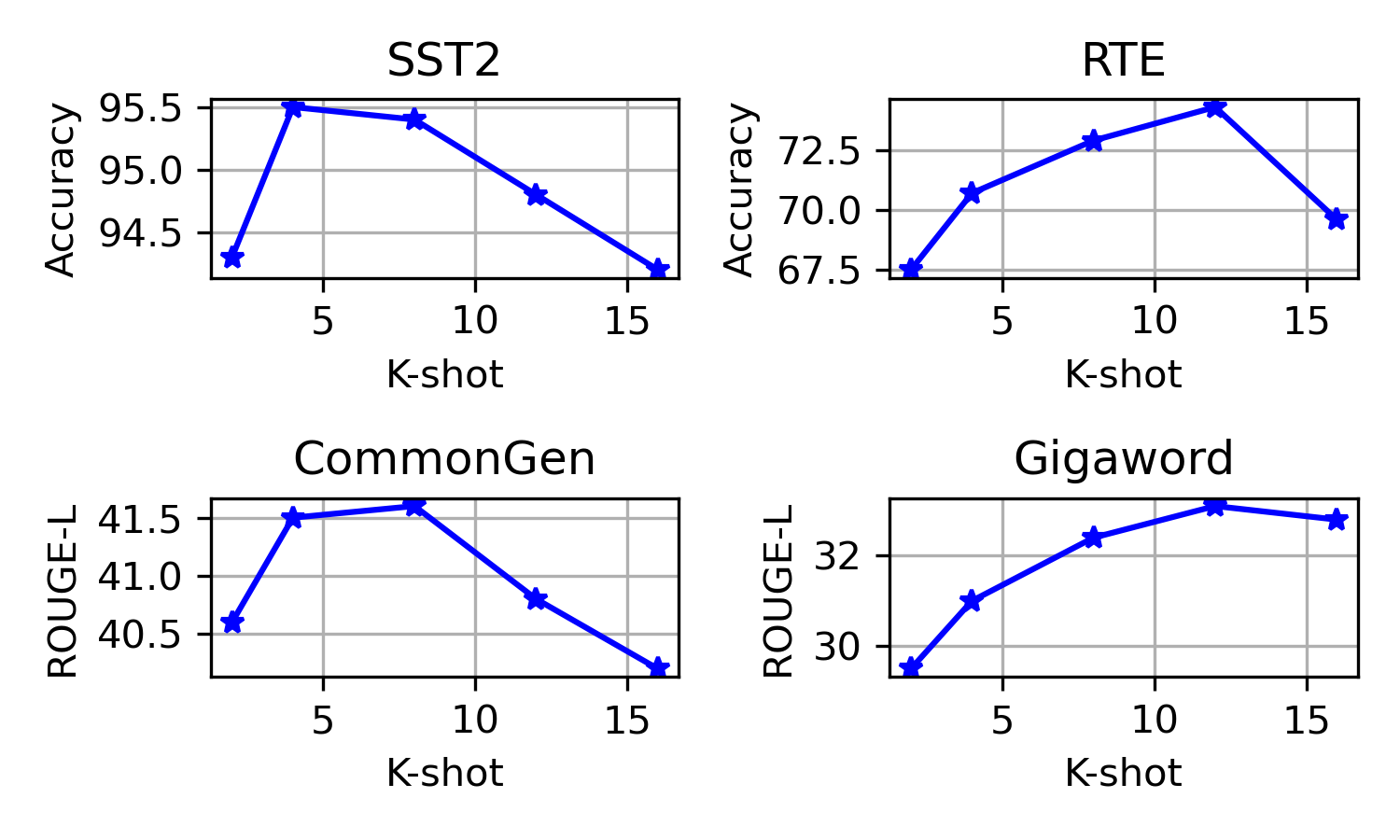} 
  \caption{The effect of demonstration number $K$.}
  \label{demonstration number}
\end{figure}

In Figure \ref{demonstration number}, we examine how the performance of \method changes as the number of demonstrations varies on the following four datasets: RTE, SST2, CommonGen, and Gigaword. 
The results show that increasing the number of demonstrations does not always lead to better performance. For SST2 and CommonGen, a significant drop in performance is observed when the number of demonstrations is too large, while Gigaword experiences a slight decrease. 

To investigate the impact of the order of demonstrations on downstream task performance, we compare three different order settings:
\begin{itemize}[leftmargin=3mm]
    \setlength{\itemsep}{0pt}
    \item Shuffle: the top-$K$ selected demonstrations are randomly shuffled. 
    \item Descending: the top-$K$ selected demonstrations are ordered in descending order of Eq.~\eqref{Demonstration Selection}.
    \item Ascending: the top-$K$ selected demonstrations are ordered in ascending order of Eq.~\eqref{Demonstration Selection}.
\end{itemize}

 The results are shown in Table~\ref{Demonstration Orders}. 
 \citet{lu2021fantastically} have revealed that ICL is sensitive to the order of demonstrations when using random examples. 
 However, we found that the order of the demonstrations selected by our method has little impact on the final in-context learning performance. 
 This supports the conclusion that high-quality demonstrations are less sensitive to the ordering \citep{li2023unified,chen2022relation,li2023mot}. 
 This further illustrates that the demonstrations selected by our method, based on preference learning, are of high quality.

\begin{table}[h]
    \centering
    \small
    \begin{tabular}{lcccc}
        \toprule
        Demo. order & Agnews & QQP & Gigaword & E2E NLG \\ 
        \midrule
        Shuffle                 & 92.8 & 81.9 & 32.6 & 55.2 \\ 
        Descending              & 92.6 & 82.0 & 33.0 & 55.1 \\ 
        Ascending               & 92.8 & 82.5 & 32.6 & 54.9 \\ 
        \bottomrule
    \end{tabular}
    \caption{Comparison of different demonstration orders.}
    \label{Demonstration Orders}
\end{table}

\section{Conclusion}\label{sec:conclusion}
In this paper, we introduced \method, a novel generative preference learning framework for optimizing demonstration selection in in-context learning (ICL). 
By reformulating ICL as a generative Bayesian optimization problem, \method bridges demonstration selection and LLM inference through a latent demonstration variable, aligning the intrinsic goal of ICL.
\method also leverages preference learning on relatively effective and ineffective demonstrations from LLM feedback, achieving fine-grained demonstration selection.
A wide range of experiments have illustrated the superiority of our proposed method \method.
We believe that our findings contribute to advancing the field of ICL and hope that \method serves as a foundation for future research in directly optimizing demonstration selection for LLMs.

\section*{Limitations}\label{sec:limit}
During the optimization phase, we treat each demonstration independently, ignoring the interactions between demonstrations (such as order and combination). In the inference phase, we select a set of demonstrations based on their scores, but the combination of individually optimal demonstrations does not necessarily result in the overall best combination.

Our method requires scoring each candidate in the demonstration pool during the demonstration selection stage. However, the search space and computational cost of this process are prohibitive. To address this, we employ E5\textsubscript{base} to reduce the search space by obtaining a subset of the demonstration pool as a new, smaller pool. However, this approach may filter out some valuable demonstrations. Additionally, improving the efficiency of demonstration scoring remains a promising direction for further exploration.

\section*{Acknowledgments}
The research received partial support from National Natural Science Foundation of China (Grant No. 62406193).
The authors also gratefully acknowledge further assistance provided by Shanghai Frontiers Science Center of Human-centered Artificial Intelligence, MoE Key Lab
of Intelligent Perception and Human-Machine Collaboration, and HPC Platform of ShanghaiTech University.

\bibliography{anthology,custom}
\bibliographystyle{acl_natbib}

\newpage
\appendix

\section{Implementation Details}
\label{Implementation Details}
We use LoRA \citep{hu2021lora} to train the parameters corresponding to the latent variable, and the total number of trainable parameters is approximately 79M.
The hyperparameters of \method are summarized in Table~\ref{tab:hyperparams}.
For a fair comparison, our method and all baselines use the same template, as shown in Appendix \ref{Templates}.

\begin{table}[h]
    \centering
    \renewcommand{\arraystretch}{1.2}
    \begin{tabular}{p{4cm} c}
        \toprule
         & \textbf{\method} \\
        \midrule
        Optimizer & AdamW \\
        Warmup steps & 3000 \\
        Training steps & 20000 \\
        Learning rate & $5\text{e-}6$ \\
        Batch size & 32 \\
        Maximum input length & 1024 \\
        Maximum output length & 64 \\
        $\beta$        & 0.1 \\
        $\lambda_w$    & 1.0 \\
        $\lambda_l$    & 1.0 \\
        \bottomrule
    \end{tabular}
    \caption{Hyper-parameters.}
    \label{tab:hyperparams}
\end{table}

\paragraph{Implementation Details of CBDS \citep{wang2024large}.}
For the \href{https://github.com/WANGXinyiLinda/concept-based-demonstration-selection}{CBDS} baseline, we used the authors' publicly available code and followed the training methodology described in the original paper. 
The original CBDS setup involves concurrent training across multiple tasks, with each task optimizing its own set of concept tokens $\theta$. 
Specifically, the optimization objective during training is to minimize  $-\log P(Y|\theta, X)$. During testing, demonstrations $(X_k, Y_k)$ are selected based on $P(\theta | X_k, Y_k)$.
Training consisted of 10,000 steps with a learning rate of 1e-2 and a batch size of 16. 
A crucial difference between CBDS and both our method and other retriever-based approaches is that CBDS uses a shared set of demonstrations for all queries within a specific task. 
This design, while computationally efficient, often leads to performance inferior to that of retriever-based methods, which dynamically select demonstrations for each query. 
This performance difference was also observed in our reproduction experiments.

\paragraph{Implementation Details of LLM-R \citep{wang2023learning}.}
Following the \href{https://huggingface.co/intfloat/llm-retriever-base/tree/main}{LLM-R} baseline procedure, we first trained a reward model and then used it for distillation to train the LLM-R model, which uses the pretrained E5\textsubscript{base} model as the initialization model. 
We utilized the code provided by the authors, retraining both the reward model and the LLM-R model while maintaining the hyperparameters from the original publication.
For EPR \citep{rubin2021learning}, we directly adopt the results presented in the LLM-R paper \citep{wang2023learning}.

\section{Details about useful samples statistics}

Figure \ref{fig:acc_distribution} illustrates the distribution of the ratio of useful examples within demonstration pools across the AGNews, HellaSwag, and OpenBookQA datasets. We first identified "hard queries" defined as those queries that remained answered incorrectly by the LLM even when provided with the top-8 demonstrations initially retrieved by the E5\textsubscript{base}~\citep{wang2022text}. Subsequently, for each identified hard query, a set of 100 demonstrations was randomly sampled without replacement from the respective dataset's full demonstration pool.

Each sampled demonstration was then individually assessed for its usefulness. A demonstration was considered "useful" if its concatenation with the hard query resulted in the LLM producing the correct answer. The \textit{Ratio of Useful Examples} was calculated for each query as $(\text{Number of Useful Demonstrations} / 100) \times 100\%$. To visualize the distribution, these ratios were binned, and the \textit{Percentage of Queries} falling within each bin was calculated as (Number of Queries in Bin $/$ Total Number of Hard Queries) $\times$ 100\%, where "Number of Queries in Bin" is the count of hard queries with a ratio within that bin's range, and "Total Number of Hard Queries" represents the total number of hard queries for that dataset.

\section{The Limitations of Contrastive Learning}
\label{metric learning}
In in-context learning, it is necessary to find demonstrations that are truly useful for the LLM to predict the ground truth. The goal of in-context learning is as follows:
\begin{equation*}
\underset{(x_k, y_k) \in \mathcal{P}}{\arg\max} P_{\mathcal{M}} (y \mid \left\{ (x_k, y_k) \right\}_{k=1}^K, x)
\end{equation*}
However, the objective of contrastive learning is to minimize the following loss.
\begin{equation*}
    \mathcal{L} = - \log \left( \frac{e^{s(x, x^+, y^+)}}{e^{s(x, (x^+, y^+))} + \sum_{i=1}^{N_{\text{neg}}} e^{s(x, (x_i^-, y_i^-))} } \right)
\end{equation*}

In the optimization process of contrastive learning, the objective shifts to optimizing the retriever by comparing the similarity between different samples. Specifically, the model is trained to distinguish between positive samples $(x^+, y^+)$ and negative samples $(x^-, y^-)$, optimizing the model by maximizing the separation between positive and negative samples. This method enlarges the distance between positive and negative samples in the embedding space, which is not aligned with in-context learning. Furthermore, it completely ignores the information about the ground truth during optimization, leading to demonstrations selected by this method that do not improve the LLM’s ability to predict the ground truth.

\section{Derivation of the Optimization Objective}
\label{Derivation of the Optimization Objective}

Starting from the marginal likelihood,
\begin{multline*} 
\log P_{\mathcal{M}}(Y \mid \{(X_k, Y_k)\}_{k=1}^K, X) = \\
\log \int_{z} P_{\mathcal{M}}(Y \mid z, X)\, \times \\
P_{\mathcal{M}}(z \mid \{(X_k, Y_k)\}_{k=1}^K, X)\, dz
\end{multline*}
we introduce an arbitrary auxiliary distribution \(q(z)\) (with \(\int q(z)\, dz = 1\)) and apply Jensen’s inequality:

\begin{align*}
\small
&\log P_{\mathcal{M}}(Y \mid \{(X_k, Y_k)\}_{k=1}^K, X)  \\
&= \log \int_{z} q(z) \times \\
&\frac{P_{\mathcal{M}}(Y \mid z, X)\, P_{\mathcal{M}}(z \mid \{(X_k, Y_k)\}_{k=1}^K, X)}{q(z)}\, dz \\[1mm] 
&\geq \int_{z} q(z) \times \\
&\log \frac{P_{\mathcal{M}}(Y \mid z, X)\, P_{\mathcal{M}}(z \mid \{(X_k, Y_k)\}_{k=1}^K, X)}{q(z)}\, dz \\[1mm]
&\triangleq \mathcal{L}(q).
\end{align*}

The quantity \(\mathcal{L}(q)\) is the evidence lower bound (ELBO) \citep{kingma2013auto}. We approximate \(q(z)\) by a Dirac \(\delta\) distribution: $q(z) = \delta(z - z^*)$ which essentially assumes that the posterior \(P_{\mathcal{M}}(z \mid \{(X_k, Y_k)\}_{k=1}^K, X)\) is highly concentrated at the optimal latent variable \(z^*\). By the sifting property of the Dirac \(\delta\),
\begin{equation*}
\int_{z} \delta(z-z^*) f(z)\, dz = f(z^*),
\end{equation*}
for any well-behaved function \(f(z)\).

Substituting this into the ELBO yields

\begin{multline*}
\mathcal{L}(q)
=\int_{z} \delta(z-z^*) \times \\
\log \frac{P_{\mathcal{M}}(Y \mid z, X)\, P_{\mathcal{M}}(z \mid \{(X_k, Y_k)\}_{k=1}^K, X)}{\delta(z-z^*)}\, dz\\[1mm]
=\log \frac{P_{\mathcal{M}}(Y \mid z^*, X)\, P_{\mathcal{M}}(z^* \mid \{(X_k, Y_k)\}_{k=1}^K, X)}{\delta(0)}.
\end{multline*}

Although \(\delta(0)\) is formally an infinite constant, it does not depend on \(z^*\) and can therefore be ignored during optimization. Maximizing the ELBO then amounts to optimizing
\begin{multline*}
\mathcal{L}(z) = -\log P_{\mathcal{M}}(Y \mid z^*, X) \\
- \log P_{\mathcal{M}}(z^* \mid \{(X_k, Y_k)\}_{k=1}^K, X)
\end{multline*}

\onecolumn
\section{Dataset Details}
\label{Templates}

\begin{table*}[h]
\centering
\small
\begin{tabular}{l|p{6cm}|p{1.7cm}|c|c|c}
\toprule
\textbf{Dataset Name} & \textbf{Template} & \textbf{Answer} & \textbf{Train} & \textbf{Test} & \textbf{Metric}  \\
\midrule
AGNews & ""\{Sentence\}" What is this text about? World, Sports, Business, or Technology?", "\{Answer\}" & 'World', 'Sports', 'Business', 'Technology'  & 120,000 & 7,600 & Accuracy  \\
\midrule
BoolQ & "\{Sentence1\} Can we conclude that \{Sentence2\}?", "\{Answer\}" & 'No', 'Yes'  & 9,427 & 3,270 & Accuracy  \\
\midrule
MultiRC & "\{Sentence1\} Question: "\{Sentence2\}" Response: "\{Sentence3\}" Does the response correctly answer the question?", "\{Answer\}"  & 'No', 'Yes'  & 27,243 & 4,848 & F1  \\
\midrule
RTE & "\{Sentence1\} Based on the paragraph above can we conclude that "\{Sentence2\}"? Yes or No?","{Answer}" & 'Yes', 'No'  & 2,490 & 277 & Accuracy  \\
\midrule
SNLI  & "If "\{Sentence1\}", does this mean that "\{Sentence2\}"? Yes, No, or Maybe?", "\{Answer\}" & 'Yes', 'Maybe', 'No' & 549,367 & 9,824 & Accuracy  \\
\midrule
PAWS & "\{Sentence1\} \{Sentence2\} Do these sentences mean the same thing?", "\{Answer\}" & 'No', 'Yes'  & 49,401 & 8,000 & Accuracy  \\
\midrule
QQP  & ""\{Sentence1\}" "\{Sentence2\}" Would you say that these questions are the same?", "\{Answer\}" & 'No', 'Yes' & 363,846 & 40,430 & Accuracy \\
\midrule
Sentiment140 & "\{Sentence\} What is the sentiment of this tweet?", "\{Answer\}" & 'Negative', 'Positive'  & 1,600,000 & 359 & Accuracy  \\
\midrule
SST2 & "Review: "\{Sentence\}" Is this movie review sentence negative or positive?", "{answer}" & 'Negative', 'Positive'  & 67,349 & 872 & Accuracy  \\
\bottomrule
\end{tabular}
\caption{Classification Task Dataset Details.}
\label{Classification Task Dataset Details}
\end{table*}

\begin{table*}[h]
\centering
\small
\begin{tabular}{l|p{6cm}|p{1.7cm}|c|c|c}
\toprule
\textbf{Dataset Name} & \textbf{Template} & \textbf{Answer} & \textbf{Train} & \textbf{Test} & \textbf{Metric}  \\
\midrule
Winogrande & "How does the sentence end? \{Sentence\}", "\{Answer\}" & 'A', 'B'  & 40,398 & 1,267 & Accuracy \\
\midrule
OpenBookQA & "\{Sentence1\} \{Sentence2\}", "\{Answer\}" & 'A', 'B', 'C', 'D'   & 4,957 & 500 & Accuracy  \\
\midrule
COPA & ""\{Sentence1\}" What is the \{Sentence2\}?", "\{Answer\}" & 'A', 'B' & 400 & 100 & Accuracy  \\
\midrule
HellaSwag & "What happens next in this paragraph? \{Sentence\}", "\{Answer\}" & 'A', 'B', 'C', 'D'  & 39,905 & 10,042 & Accuracy  \\
\bottomrule
\end{tabular}
\caption{Multi-choice Task Dataset Details.}
\label{Multi-choice Task Dataset Details}
\end{table*}

\begin{table*}[h]
\centering
\small
\begin{tabular}{l|p{6cm}|c|c|c}
\toprule
\textbf{Dataset Name} & \textbf{Template} & \textbf{Train} & \textbf{Test} & \textbf{Metric}  \\
\midrule
AESLC & "What is the subject line for this email? \{Sentence\}", "\{Label\}"   & 13,181 & 1,750 & ROUGE-L \\
\midrule
Gigaword  & "Write a short summary for this text: \{Sentence\}", "\{Answer\}" & 2,044,465 & 730 & ROUGE-L  \\
\midrule
SQuAD v1 & "Please answer a question about the following article about \{Sentence\}: \{Sentence1\} \{Sentence2\}", "\{Answer\}"  & 87,599 & 10,570 & Exact Match  \\
\midrule
CommonGen & "Concepts: \{Sentence\}. Write a sentence that includes all these words.", "\{Answer\}"  & 67,389 & 4,018 & ROUGE-L  \\
\midrule
DART & "Triple: \{Sentence\} What is a sentence that describes this triple?", "\{Answer\}"  & 62,659 & 2,768 & ROUGE-L  \\
\midrule
E2E NLG & "Attributes: \{Sentence\}. Produce a detailed sentence about this restaurant.", "\{Answer\}"  & 33,525 & 1,847 & ROUGE-L  \\

\bottomrule
\end{tabular}
\caption{Generation Task Dataset Details.}
\label{Generation Task Dataset Details}
\end{table*}

\newpage
\section{Case Study of Our \method Method}
\label{appendix:case_study}

\begin{table*}[h]
\centering
\begin{tabular}{@{}ll@{}}
\toprule
\textbf{Task name} & AGNews \\
\textbf{Test Input} & \begin{minipage}[t]{0.75\textwidth}
"Dominant US captures gold with 79th straight win The US softball team completed its scorched-earth run through the Olympics on Monday with a 5-1 win over Australia, America's third straight gold medal." What is this text about? World, Sports, Business, or Technology? \\
\end{minipage} \\
\textbf{Demonstration} & \begin{minipage}[t]{0.75\textwidth}
"US Women Shatter Olympic 800-Meter Freestyle Relay Record The United States has shattered a 17-year-old world record in the women's Olympic 800-meter freestyle relay. " What is this text about? World, Sports, Business, or Technology? Sports \\
\end{minipage} \\
\textbf{Test Answer} & \begin{minipage}[t]{0.75\textwidth}
"Sports"
\end{minipage} \\
\midrule
\textbf{Task name} & Sentiment140 \\
\textbf{Test Input} & \begin{minipage}[t]{0.75\textwidth}
zomg!!! I have a G2!!!!!!! What is the sentiment of this tweet?
\end{minipage} \\
\textbf{Demonstration} & \begin{minipage}[t]{0.75\textwidth}
My brother got his update for his G1 and I ain't got shit  What is the sentiment of this tweet? Negative
\end{minipage} \\
\textbf{Test Answer} & \begin{minipage}[t]{0.75\textwidth}
"Positive"
\end{minipage} \\
\midrule
\textbf{Task name} & RTE \\
\textbf{Test Input} & \begin{minipage}[t]{0.75\textwidth}
The west has preferred to focus on endangered animals, rather than endangered humans. African elephants are hunted down and stripped of tusks and hidden by poachers. Their numbers in Africa slumped from 1.2m to 600,000 in a decade until CITES - the Convention on International Trade in Endangered Species - banned the trade in ivory. Based on the paragraph above can we conclude that "African elephants are endangered by ivory poachers."? Yes or No?
\end{minipage} \\
\textbf{Demonstration} & \begin{minipage}[t]{0.75\textwidth}
Three leading Japanese banks have announced an alliance forming the world's largest financial group. Fuji Bank, Dai-Ichi Kangyo and the Industrial Bank of Japan say their operations will be integrated by the spring of 2002. Based on the paragraph above can we conclude that "Merger of Japanese Banks creates the world's biggest bank."? Yes or No? Yes
\end{minipage} \\
\textbf{Test Answer} & \begin{minipage}[t]{0.75\textwidth}
"Yes"
\end{minipage} \\
\midrule
\textbf{Task name} & SST2 \\
\textbf{Test Input} & \begin{minipage}[t]{0.75\textwidth}
Review: "instead of a hyperbolic beat-charged urban western , it 's an unpretentious , sociologically pointed slice of life . " Is this movie review sentence negative or positive?
\end{minipage} \\
\textbf{Demonstration} & \begin{minipage}[t]{0.75\textwidth}
Review: "offers an unexpected window into the complexities of the middle east struggle and into the humanity of its people . " Is this movie review sentence negative or positive? Positive
\end{minipage} \\
\textbf{Test Answer} & \begin{minipage}[t]{0.75\textwidth}
"Positive"
\end{minipage} \\
\midrule
\textbf{Task name} & CommonGen \\
\textbf{Test Input} & \begin{minipage}[t]{0.75\textwidth}
Concepts: kid, yard, ball. Write a sentence that includes all these words.
\end{minipage} \\
\textbf{Demonstration} & \begin{minipage}[t]{0.75\textwidth}
Concepts: kid, grass, crawl. Write a sentence that includes all these words. A kid is about to crawl through some grass.
\end{minipage} \\
\textbf{Test Answer} & \begin{minipage}[t]{0.75\textwidth}
"A kid is playing with a ball in his yard."
\end{minipage} \\

\bottomrule
\end{tabular}
\caption{Case Study of Our \method.}
\label{tab:case_study}
\end{table*}

\end{document}